\documentclass[lettersize,journal]{IEEEtran}
\usepackage{amsmath,amsfonts}
\usepackage{algorithmic}
\usepackage{algorithm}
\usepackage{array}
\usepackage[caption=false,font=normalsize,labelfont=sf,textfont=sf]{subfig}
\usepackage{textcomp}
\usepackage{stfloats}
\usepackage[flushleft]{threeparttable}
\usepackage{url}
\usepackage{verbatim}
\usepackage{graphicx}
\usepackage{cite}
\usepackage[T1]{fontenc}
\usepackage{aecompl}
\begin{document}

\title{Light-LOAM: A Lightweight LiDAR Odometry and Mapping based on Graph-Matching}

\author{Shiquan Yi, Yang Lyu~\IEEEmembership{Member,~IEEE}, Lin Hua, Quan Pan~\IEEEmembership{Member,~IEEE}, Chunhui Zhao~\IEEEmembership{Member,~IEEE}
	\thanks{Shiquan Yi, Yang Lyu, Lin Hua, Quan Pan, Chunhui Zhao are with the School of Automation, Northwestern Polytechnical University, Xi'an Shaanxi, 710072 China. email: lyu.yang@nwpu.edu.cn}
}


\IEEEpubid{0000--0000/00\$00.00~\copyright~2021 IEEE}

\maketitle
\pagestyle{empty}  
\thispagestyle{empty} 
\begin{abstract}
Simultaneous Localization and Mapping (SLAM) plays an important role in robot autonomy. Reliability and efficiency are the two most valued features for applying SLAM in robot applications. In this paper, we consider achieving a reliable LiDAR-based SLAM function in computation-limited platforms, such as quadrotor UAVs based on graph-based point cloud association.
First, contrary to most works selecting salient features for point cloud registration, we propose a non-conspicuous feature selection strategy for reliability and robustness purposes. Then a two-stage correspondence selection method is used to register the point cloud, which includes a KD-tree-based coarse matching followed by a graph-based matching method that uses geometric consistency to vote out incorrect correspondences.
Additionally, we propose an odometry approach where the weight optimizations are guided by vote results from the aforementioned geometric consistency graph. In this way, the optimization of LiDAR odometry rapidly converges and evaluates a fairly accurate transformation resulting in the back-end module efficiently finishing the mapping task. Finally, we evaluate our proposed framework on the KITTI odometry dataset and real-world environments. Experiments show that our SLAM system achieves a comparative level or higher level of accuracy with more balanced computation efficiency compared with the mainstream LiDAR-based SLAM solutions. 
\end{abstract}


\section{Introduction}
Simultaneously Localization and Mapping (SLAM) is now considered one indispensable module for many mobile robots to achieve autonomous navigation in challenging GNSS-denied environments. 
To adapt to different missions and platforms, various SLAM frameworks based on different sensors, processing methods, or functional structures have been proposed. Disregarding the taxonomy, most SLAM works focus on improving two indices, which are 1) localization performance, such as accuracy and reliability, and 2) runtime efficiency, such as computation burden and storage requirement. 
Usually, the two indices cannot be optimized both on a resource-limited robot platform. Rather, a trade-off between performance and efficiency is typically required to set up a proper SLAM system function in practical robot applications. In this paper, we aim to develop a lightweight LiDAR-based SLAM with comparative performance to the state-of-the-art methods, but with balanced computation requirements so as to adapt to resource-limited platforms.
\begin{figure}[h]
	\centering
	\includegraphics[width=0.48\textwidth]{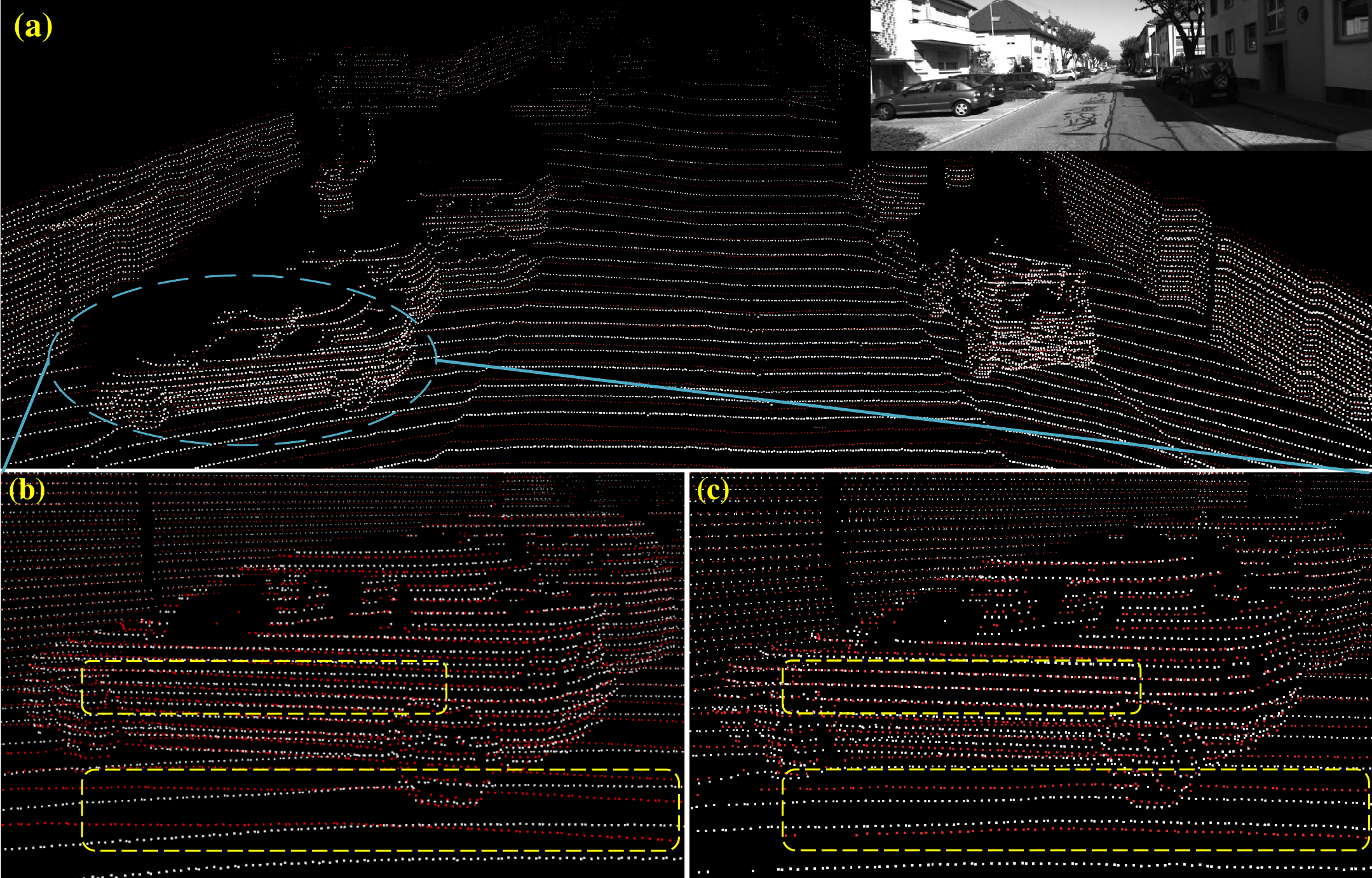}
	\caption{Feature points alignment between two consecutive scans using different data association methods. (a) A scan of a scenario's point cloud. (b) K-nearest neighbor method. (c) Graph-based two-stage matching method.}
	\label{fig:comparison}
\end{figure}
Vision-based and LiDAR-based SLAMs are the two most widely researched streamlines according to the sensor types. Vision-based methods provide the most economical and practical localization solutions in many stable environments.
On the other hand, LiDAR-based methods are usually preferred in structure-rich environments and are considered to be more robust against light condition changes than vision sensors. There are several mile-stone LiDAR-based works, such as LiDAR Odometry and Mapping (LOAM)\cite{zhang2014loam}, lightweight and ground-optimized LiDAR odometry and mapping (LeGO-LOAM)\cite{shan2018lego} and Cartographer \cite{hess2016real}. 
Roughly, the LiDAR-base SLAM function can be divided into two subfunctions, which are the front-end odometry, and the back-end map optimization. In the front-end part, the main purpose is to obtain the incremental transformations between consecutive frames through scan-to-scan or scan-to-map alignment. In the back-end part, an estimator/optimizer that integrates more state constraints is utilized to improve the front-end estimation accuracy and smoothness. 
In most SLAM frameworks, the back-end task consumes more computation resources than the front-end and is usually updated more slowly. With the utilization of the multi-thread technique, the slower thread determines the functional frequency of the whole SLAM function. 
In a robot control loop, a slower map update rate may affect the robot feedback reaction in unknown environments. 
Therefore, we considered achieving a more balanced SLAM functional partition with improved performance for practical robot applications. More specifically, we put more computation effort in the front end to improve the point cloud alignment performance, and thanks to a more accurate front end, we can shrink the computation complexity of the back end part.

In the front-end part, there are mainly two streams to achieve the point cloud registration between scans or between scan and map. The first stream is to obtain relative transformation directly from point-to-point correspondences. Iterative closest point (ICP) and its variants, such as the GICP \cite{segal2009generalized}, are the most widely used methods to align two-point clouds. ICP-based methods are widely used in early-stage 2D LiDAR-based SLAM frameworks when the size of the point cloud is much more sparse than that of 3D LiDARs.
Besides, the Normal distribution transformation (NDT) is considered another direct point cloud registration method. Although it uses all points, the NDT method represents the point cloud with normal distributions and then calculates the relative transformations in a probabilistic fashion. 
However, the direct methods use all points to optimize the relative transformation, therefore its computation requirement can be a major concern for processing real-time 3D LiDAR scan sequences. The second stream is to extract salient features from point clouds, and then to carry out registration only with the feature points. In the milestone work, LOAM, points are selected based on their curvature and then assigned as planar points and edge points, and then registration is carried out based on the nearest neighborhood methods. Similar geometric features are also used in LeGO-LOAM\cite{shan2018lego} and F-LOAM\cite{flaom}. Besides the geometric calculated features, there is also a trend to implement learning-based methods to obtain features represented as deep neural networks due to their power to represent nonlinearity in the descriptors. However, the learning-based methods are often criticized as data-dependent and may not be ready to be used in unknown environments. 
Although with different representations of the point cloud scans, most methods above use KD-tree-based technology for efficient correspondence indexing. While KD-tree\cite{muja2009fast} is widely used for establishing initial correspondences in SLAM systems\cite{zhang2014loam,flaom,shan2018lego,shan2020lio,xu2021fast}, it can introduce erroneous associations, particularly in noisy or occluded environments.

Recent years have seen a notable increase in research efforts directed toward leveraging the principles of geometric consistency and graph theory to tackle the point cloud data association problems.
Bailey et al.'s pioneering work \cite{bailey2000data} introduced the application of geometric consistency in addressing the 2D LiDAR-based map-building problem. This approach involves the construction of a graph, wherein the selection of the correct correspondence set is facilitated through the identification of the maximum common subgraph.
In a related vein, Lajoie et al. \cite{Lajoie_Ramtoula_Chang_Carlone_Beltrame_2020} utilized pairwise consistent measurements to mitigate spurious loop closures within a distributed SLAM system. This methodology similarly involves the maximization of a consistent subgraph, which finds application in the domain of multi-robot map merging, as documented in the literature \cite{Mangelson_Dominic_Eustice_Vasudevan_2018}.
Yang et al.\cite{Yang_Shi_Carlone_2021} introduced the TEASER graph-theoretic framework, which incorporates the truncated least squares optimization method and maximum clique inlier selection technology to effectively eliminate numerous spurious correspondences. Its efficacy has been substantiated in point registration and scan-matching tasks. 
In the studies \cite{yang2022correspondence} and \cite{yang2023mutual}, the researchers utilize geometrically consistent graphs in conjunction with a variety of voting strategies to rank correspondences and select dependable inliers.
Another graph-theoretic framework development, named Clipper \cite{lusk2021clipper},  constructs a weighted graph and formulates the inlier association as an optimization problem, ultimately solving for the densest subgraph of consistent correspondences. While these graph-based solutions for association problems have demonstrated effectiveness, particularly in adhering to the geometric consistency constraint, it remains a challenge to optimize their efficiency. Notably, their performance falls short of desired levels when confronted with large-scale point cloud alignment within stringent time constraints, for example, in a LiDAR-based SLAM framework.

In this paper, we consider achieving a reliable LiDAR-based SLAM function in computation-limited robot platforms. Contribution are mainly on two folds. First, 
we develop an innovative SLAM front-end which includes a non-conspicious feature selection strategy and a graph-based feature matching function to achieve better point cloud registration. Secondly, to benefit from the reliable registration of the front end, we develop a light backend that can be executed more efficiently on a computation-limited platform. Experiments validations are carried out with both public datasets and self-collected data. Our implementation of Light-LOAM will be freely available at: \url{https://github.com/BrenYi/Light-LOAM}.

The remainder of the paper is organized as follows. In Section II, we give a description of the proposed method, and Section IV provides the experimental results. Section V concludes the paper.

\begin{figure*}[t]
	
	\centering
	\includegraphics[width=0.9\textwidth]{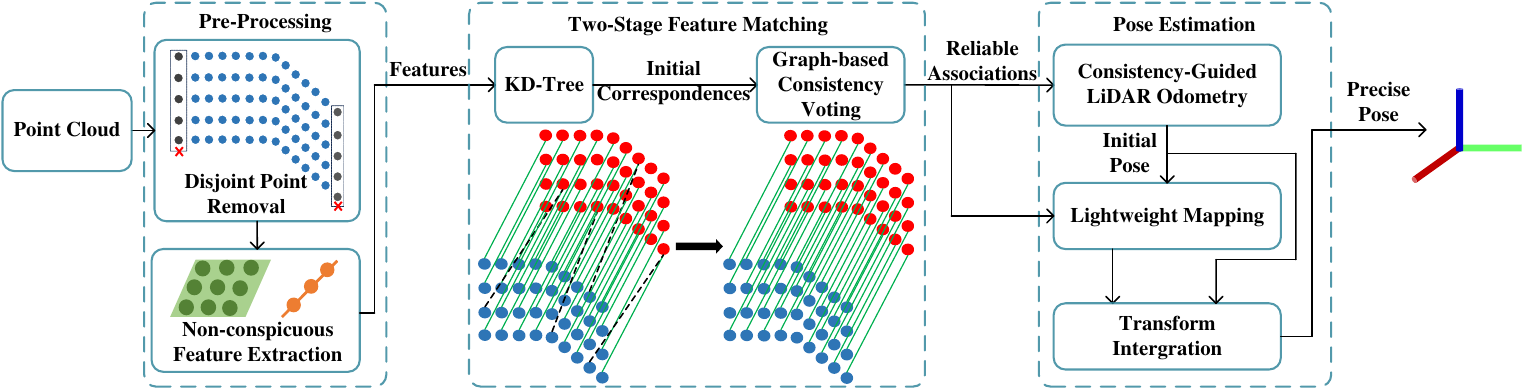}
	\caption{Overview of Light-LOAM System}
	\label{fig:overview_of_Light-LOAM_system}
\end{figure*}
\begin{figure*}[t]
	\centering
	\includegraphics[width=0.95\textwidth]{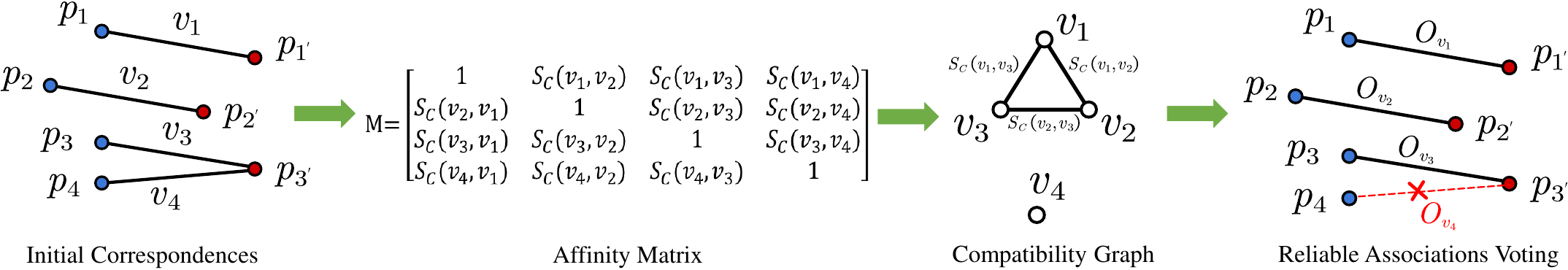}
	\caption{Process of Graph-based Two-stage Feature Matching}
	\label{fig:overview_of_graph-based_matching}
\end{figure*}

\section{System Overview}
We present the pipeline of our Light-LOAM SLAM system in Fig. \ref{fig:overview_of_Light-LOAM_system}, which is composed of three core stages: {\it{pre-processing}}, {\it{two-stage feature matching}}, and {\it pose estimation}. 
In the pre-processing stage, we begin by filtering out disjoint points from each point cloud scan. To select stable corner and plane features with subtle local geometry attributes, we employ a non-conspicuous selection method and filter out the most significant corner and plane features. This is one main difference against other methods\cite{zhang2014loam,shan2018lego,flaom}.
A two-stage feature-matching process is carried out afterward. In the first stage, a KD-Tree-based method\cite{zhang2014loam} is employed to establish initial correspondences for the selected features. Then, we introduce a graph-based consistency voting mechanism to assess these correspondence relationships, effectively filtering out unreliable associations.
Moving to the front-end odometry module, the consistent scores of reliable point pairs are leveraged to optimize transformations, resulting in initial, relatively precise pose estimations.
Finally, with the support of these initial reliable estimations,  the mapping module optimizes more accurate poses in a more efficient fashion.

\subsection{Feature Extraction and Selection}
\paragraph{Disjoint Point Removal} Given the substantial volume of data produced by 3D LiDAR sensors, feature extraction and feature-based alignment is a widely adopted approach for efficient transformation evaluation. However, it is imperative to eliminate disjoint objects before extracting feature candidates. Disjoint points may often represent outliers or segments of occluded objects, and their inclusion can significantly degrade subsequent feature association and pose estimation quality. Therefore, in alignment with prior work \cite{disjoint_paper}, we employ the following criteria to exclude these discontinued points:
\begin{equation}
	\left | \left \| \mathbf{p}_{i+1}^{k}  - \mathbf{p}_{i}^{k} \right \|_{2} -   \left \| \mathbf{p}_{i-1}^{k}  -  \mathbf{p}_{i}^{k}\right \|_{2}  \right | > \sigma _{disjoint}
	\label{euqal:disconnectivity}
\end{equation}
where $\mathbf{p}_{i}^{k}$ represents point $i$ located in the $k$th laser beam channel, and $\sigma _{disjoint}$ serves as the judgment threshold. A point is classified as disjoint if the absolute difference between its Euclidean distances to neighbors on both sides exceeds $\sigma _{disjoint}$. Otherwise, it is considered a consecutive candidate.

After eliminating disjoint objects, we extract feature points from each laser beam channel. The local geometric attribute of a point is characterized using the smoothness metric (\ref{equation::roughness}).
\begin{equation}
	r(\mathbf{p}_{i}^{k})=\frac{1}{\left | \mathit{S} \right |\left \| \mathbf{p}_{i}^{k} \right \|}\left \| \sum_{j\in  S,j\notin i }(\mathbf{p}_{i}^{k}-\mathbf{p}_{j}^{k})\right\| 
	\label{equation::roughness} 
\end{equation}
where $\mathit{S}$ denotes the assessed point set, encompassing the candidate and its adjacent objects on both the left and right sides. A candidate is designated as a corner feature if its smoothness $r_{i}^{k}$ exceeds the threshold $r_{t}$; otherwise, it is identified as a plane feature. This classification process involves considering 5 points on both sides of a candidate, and the threshold $r_{t}$ is set to $0.1$ for practical implementation.

In conventional LiDAR-based SLAM systems, such as LOAM\cite{zhang2014loam}, FLOAM\cite{flaom}, LEGO-LOAM\cite{shan2018lego}, the perceived space is commonly divided into several subregions. Feature candidates with the highest or lowest smoothness attributes are selected from each subregion for subsequent feature matching.
However, our Light-LOAM SLAM system introduces an innovative non-conspicuous feature selection strategy. As mentioned earlier, feature selection is typically guided by a discriminative principle. But, are these discriminative features truly robust and capable of serving as high-quality optimization samples? It's worth noting that some outliers or occluded points can exhibit highly discriminative geometric attributes. Therefore, we hold the view that candidates with weaker smoothness attributes than the top conspicuous ones may be more valuable and robust for data association.

We prioritize the selection of weaker corner and plane features as our optimized candidates within each subregion. Before initiating the feature selection process, points are first sorted in descending order based on their smoothness values within each subregion:
\begin{equation}
	\mathbb{F}^{k} = \left \{ \mathbf{p}_{i}^{k},\cdots,\mathbf{p}_{i+j}^{k}| r(\mathbf{p}_{i}^{k}) >\cdots>r(\mathbf{p}_{i+j}^{k})\right \} 
	\label{equation:ordered_set}
\end{equation}

We choose the $m$ sharpest points after the first $k$ points in the ordered set $\mathbb{F}^{k}$ and designate them as the edge feature set $\mathbb{F}^{k}_{e}$. Similarly, we select the $n$ flattest candidates before the last $l$ points and include them in the planar set $\mathbb{F}^{k}_{s}$:
\begin{equation}
	\mathbb{F}^{k}_{e} \! = \!\left \{ \mathbf{p}_{i+k+1}^{k},\cdots,\mathbf{p}_{i+k+m}^{k}\mid \mathbf{p}_{x}^{k}\in \mathbb{F}^{k},r(\mathbf{p}_{x}^{k})>r_{t}\right \} 
\end{equation}
\begin{equation}
	\mathbb{F}^{k}_{s}\! = \! \left \{ \mathbf{p}_{i+j-l-n}^{k},\cdots,\mathbf{p}_{i+j-l-1}^{k} \mid  \mathbf{p}_{x}^{k}\in \mathbb{F}^{k},r(\mathbf{p}_{x}^{k})<r_{t} \right \}  
\end{equation}

In our implementation, we horizontally divide each laser beam channel into 6 subregions. Setting $m$, $n$, $k$, and $l$ to 2, 4, 1, and 2, respectively, this non-conspicuous feature selection method, combined with the disjoint point removal preprocessing step, efficiently filters out more outliers, ensuring a more reliable set of candidates for subsequent pose estimation.

\subsection{Graph-based Two-stage Feature Matching}
\begin{figure}[h]
	\centering
	\includegraphics[width=0.4\textwidth]{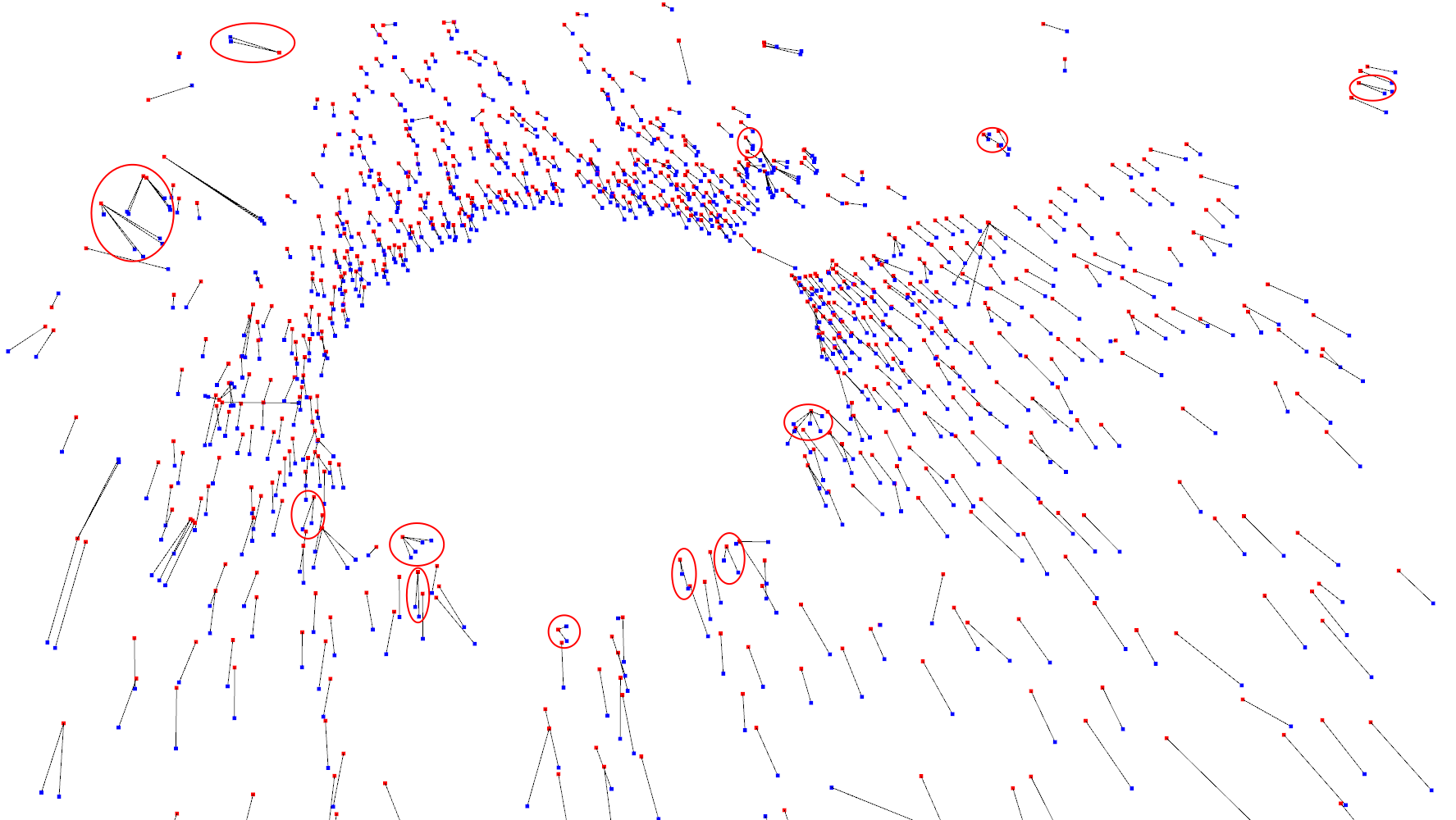}
	\caption{The demonstration of initial feature correspondences generated by KD-tree. In each correspondence, the red point is the source feature from the current scan, and the blue one is its corresponding target object from the last scan or map. Ellipses indicate incorrect data associations of one scan. }
	\label{fig:kdtree_error}
\end{figure}

Identifying corresponding features from both the last scan of the point cloud and the existing constructed map is a fundamental prerequisite for subsequent scan-to-scan and scan-to-map alignments.
KD-tree\cite{muja2009fast} is a widely used method for establishing correspondence relationships due to its efficiency and effectiveness, as evidenced by its adoption in various works\cite{zhang2014loam,shan2018lego,flaom}.
Despite its prevalence, KD-tree is susceptible to errors caused by environmental occlusions, outliers, and noise within the point cloud, leading to inaccurate pose estimations.
For instance, as shown in Fig. \ref{fig:kdtree_error}, situations can arise where more than one candidate feature from the current scan matches the same point from the last scan of the point cloud as its closest counterpart, leading to false multi-to-one correspondence cases. To mitigate such issues and reduce spurious correspondences, we introduce a novel graph-based two-stage correspondence selection method.
\paragraph{Initial Correspondences Determination by KD-tree}We start by using KD-tree to find correspondences for our feature candidates, assuming the closest point from the last scan or map is the true correspondence for each feature. This establishes our initial set of point pairs using the formula (\ref{equation::cor_relationship}).
\begin{equation}
	\varpi = \left \{ \left ( \mathbf{p}_{i},\mathbf{p}_{i^{\prime}} \right ) \mid \mathbf{p}_{i}\in \mathbb{F}^{k}_{e}\cup \mathbb{F}^{k}_{s},\mathbf{p}_{i^{\prime }}\in\mathbb{F}^{k}_{e^{\prime }} \cup \mathbb{F}^{k}_{s^{\prime }}\right \}  
	\label{equation::cor_relationship}
\end{equation}
where $\mathbf{p}{i}$ represents features from the edge set $\mathbb{F}^{k}_{e}$ or the planar group $\mathbb{F}^{k}_{s}$ of the current scan, and $\mathbf{p}_{i^{\prime}}$ corresponds to their closest corresponding point from the feature set of the last scan of the point cloud or the map.

\paragraph{Reliable Associations via Consistent Graph} In the second stage, a graph-based correspondence validation algorithm is introduced. Beginning with the initial putative associations $\varpi$ from the KD-tree, a compatibility graph is constructed based on the principle of geometric consistency.

Before delving into the concept of geometric consistency, let's assume the existence of two correct associations, $\left ( \mathbf{p}_{i},\mathbf{p}_{i^{\prime}} \right )$ and $\left ( \mathbf{p}_{j},\mathbf{p}_{j^{\prime}} \right )$, which share an identical set of transformation parameters denoted as $\left ( \mathbf{R}, \mathbf{T}\right )$. These two pairs of associations can be formulated as:
\begin{equation}
	\mathbf{p}_{i^{\prime}} = \mathbf{R}\mathbf{p}_{i}+\mathbf{T}
\end{equation}
\begin{equation}
	\mathbf{p}_{j^{\prime}} = \mathbf{R}\mathbf{p}_{j}+\mathbf{T}
\end{equation}

Theoretically, the Euclidean distance between two target points remains constant across different frames, as expressed by (\ref{eaqution:consitency}), embodying what we refer to as geometric consistency.
\begin{equation}
	\left \|\mathbf{p}_{i^{\prime}}-\mathbf{p}_{j^{\prime}}  \right \|_{2} =\left \|\mathbf{p}_{i}-\mathbf{p}_{j}\right \|_{2}
	\label{eaqution:consitency}
\end{equation}

We can leverage this constraint to evaluate the compatibility of correspondences within a graph space, rather than the Euclidean space.
To illustrate this conveniently, let's assume there are four hypothetical association cases generated from the KD-tree, as depicted in Fig. \ref{fig:overview_of_graph-based_matching}. 
We can construct a compatibility graph, where each vertex, denoted as $v_{i}=(\mathbf{p}_{i},\mathbf{p}_{i^{\prime}})$, represents the $i$th association relationship. The edges in the graph indicate that the two associations, $v_{i}$ and $v_{j}$, are compatible or geometrically consistent. Within the graph displayed in Fig. \ref{fig:overview_of_graph-based_matching}, there exist four associations: $v_{1}$, $v_{2}$, $v_{3}$, and $v_{4}$. Notably, $v_{1}$, $v_{2}$, and $v_{3}$ are mutually geometrically consistent.

Following this, as depicted in Fig. \ref{fig:overview_of_graph-based_matching}, the compatibility graph is constructed using an affinity matrix $M$. Each entry $M(i, j)$ in this matrix represents the geometric consistency score of the correspondence pair $\left(v_{i}, v_{j}\right)$ and is calculated quantitatively as:
\begin{equation}
	S_{c}(v_{i}, v_{j})=\mathrm{exp} (-\frac{d(v_{i},v_{j})^{2}}{\sigma^{2}} )
	\label{equation::score}
\end{equation}
Here, the term $d(v_{i}, v_{j})$ is defined as:
\begin{equation}
	d(v_{i},v_{j}) = \left \|\mathbf{p}_{i^{\prime}}-\mathbf{p}_{j^{\prime}}  \right \|_{2} -\left \|\mathbf{p}_{i}-\mathbf{p}_{j}\right \|_{2}
\end{equation}
where  $\sigma$ serves as a distance adjustment parameter. Notably, $S_{c}$ ranges from 0 to 1, achieving 1 for perfect geometric consistency. The diagonal entry $M(i, i)$ consistently equals 1. A lower score indicates a higher degree of inconsistency.

Based on the compatibility graph, we employ a voting rule to assess the quality of each correspondence. This voting mechanism can be expressed as:
\begin{equation}
	o_{i}=o(v_{i})= \sum_{j=0,j\ne i}^{\left | \varpi  \right | }\left \lfloor \frac{S_{c}(v_{i}, v_{j})}{\eta}  \right \rfloor 
\end{equation}
where $\eta$ functions as the voting threshold that determines the compatibility of two associations in the voting process. Additionally, $\left | \varpi \right |$ denotes the cardinality of the correspondence set $\varpi$. In this scheme, an association subject $v_{i}$ receives one vote if its consistency score $S_{c}(v_{i}, v_{j})$ with association $v_{j}$ meets or exceeds the threshold $\eta$. The final level of consistency in the voting process is determined by all consistent voters.

Following the completion of our voting pipeline, we can represent the sequence of voting results in descending order as follows:
\begin{equation}
	\mathbb{O}  =\left \{ o_{i}\mid o_{1}> o_{2}>\cdots>o_{i}, i\in \left [ 1,\left | \varpi  \right | \right ] \right \}   
\end{equation}

In the event that a correspondence candidate $v_i$ receives a voting score $o_i$ lower than $x\%$ of the total number of association candidates, it is considered an unreliable association and is subsequently filtered out. After removing these outlier associations, we obtain the final set of reliable associations along with their corresponding scores:
\begin{equation}
	\mathbb{O}^{\prime}  =\left \{ o_{i}\mid o_{i}\in \mathbb{O}, o_{i} > x\left | \mathbb{O}  \right |\right \}  
\end{equation}
\begin{equation}
	\varpi^{\prime} = \left \{ v_{i} \mid v_{i} \in \varpi, o(v_{i}) \in \mathbb{O}^{\prime}\right \}  
	\label{equation::cor_relationship1}
\end{equation}

In analyzing the computational complexity of our graph-based matching algorithm with N correspondences, the construction of the compatibility graph and the correspondence ranking using quicksort have time complexities of $O(N^{2})$ and $O(N\log{N})$, respectively. 
This results in a total time complexity of $O(N^{2}+N\log{N})\!=\!O(N^{2})$ for our graph-based voting algorithm.
To maintain real-time performance, we partition the perceptual space into n subregions. Correspondence relationships within each subregion form subgraphs, handling tasks in both odometry and mapping stages. In the odometry stage, each subregion processes around 200 correspondences, averaging a total processing time of approximately 3 ms. During the mapping stage, each subregion deals with roughly 350 correspondences, with an average total processing time of about 7 ms. This correspondence division across subregions ensures both real-time performance and accurate results.

\subsection{Consistency-Guided LiDAR Odometry}
In LiDAR SLAM systems, odometry is pivotal for refining initial poses through scan-to-scan point cloud matching. The odometry module usually provides high-frequency but somewhat imprecise pose estimations, acting as an initial input for the mapping module. More accurate initial transformations estimated by the odometry module accelerate the convergence of the final robot pose estimation, resulting in a reduction in the computational cost of the mapping back-end.In light of this, we propose a novel LiDAR odometry mechanism in which pose optimization is guided by the voting results from the compatibility graph. 

In our odometry module, we aim to optimize the transformation $T_{k}^{k-1}\in SE(3)$ representing the motion from the $k$th frame to the $(k-1)$th frame, and update the global pose $T_{k}^{W}\in SE(3)$ of the point cloud in the $k$-th frame. Prior to optimization, we correct motion distortions in the point cloud by assuming uniform motion. Similar to LOAM\cite{zhang2014loam}, we define two types of residual terms. The first term is the point-to-line distance residual, given by:
\begin{equation}
	f_{e}^{L}(v_{i})= \frac{\left |(\mathbf{\tilde{p}}_{(i,k)}^{L}-\mathbf{p}_{(i^{\prime },k-1)}^{L})\times (\mathbf{\tilde{p}} _{(i,k)}^{L}-\mathbf{p}_{(j^{\prime },k-1)}^{L})\right | }{\left |\mathbf{p}_{(i^{\prime },k-1)}^{L}-\mathbf{p}_{(j^{\prime },k-1)}^{L}  \right | } 
	\label{equation::residual_edge}
\end{equation}

In (\ref{equation::residual_edge}), $\mathbf{\tilde{p}}_{(i,k)}^{L} \!=\! T_{k}^{k-1}\mathbf{p}_{(i,k)}^{L}$, where $\mathbf{p}_{(i,k)}^{L} \! \in \! \mathbb{F}^{k}_{e}$ represents a corner feature. $\mathbf{p}_{(i^{\prime },k-1)}^{L} \in \mathbb{F}^{k-1}_{e}$  represents the closest object from the same laser beam channel to $\mathbf{\tilde{p}}_{(i,k)}^{L}$.
The point-pair $(\mathbf{p}_{(i,k)}^{L}, \mathbf{p}_{(i^{\prime },k-1)}^{L})\in \varpi^{\prime}$ signifies a reliable correspondence identified through our graph-based two-stage feature matching. Additionally, $\mathbf{p}_{(j^{\prime },k-1)}^{L} \in \mathbb{F}^{k-1}_{e}$ is another nearest neighbor of $\mathbf{\tilde{p}}_{(i,k)}^{L}$ from different laser channel. $\mathbf{p}_{(i^{\prime },k-1)}^{L}$ and $\mathbf{p}_{(j^{\prime },k-1)}^{L}$ together form a line geometric residual term.

For the planar residual, we have:
\begin{equation}
	f_{s}^{L}(v_{i})= \mathbf{n_{s}} \cdot (\mathbf{\tilde{p}}_{(i,k)}^{L}-\mathbf{p}_{(i^{\prime },k-1)}^{L})  
	\label{equation::residual_planar}
\end{equation}
where 
\begin{equation}
	\mathbf{n_{s}}\!= \!\frac{(\mathbf{p}_{(i^{\prime },k-1)}^{L}-\mathbf{p}_{(j^{\prime },k-1)}^{L})\!\times\! (\mathbf{p}_{(i^{\prime },k-1)}^{L}-\mathbf{p}_{(l^{\prime },k-1)}^{L})}{\left |(\mathbf{p}_{(i^{\prime },k-1)}^{L}-\mathbf{p}_{(j^{\prime },k-1)}^{L})\!\times\! (\mathbf{p}_{(i^{\prime },k-1)}^{L}-\mathbf{p}_{(l^{\prime },k-1)}^{L})\right | }
\end{equation}

In (\ref{equation::residual_planar}), $\mathbf{\tilde{p}}_{(i,k)}^{L} = T_{k}^{k-1}\mathbf{p}_{i,k}^{L}$ where $\mathbf{p}_{i,k}^{L} \in \mathbb{F}^{k}_{s}$ is a planar feature. $\mathbf{p}_{(i^{\prime },k-1)}^{L}$ and $\mathbf{p}_{(j^{\prime },k-1)}^{L}$ are the first and second closest objects to the projected feature $\mathbf{\tilde{p}}_{(i,k)}^{L}$ from the same laser beam channel. Additionally, $\mathbf{p}_{(l^{\prime },k-1)}^{L}$ is the other nearest point from a different channel. These three points from the $k-1$th scan collectively construct a plane to establish a point-to-plane residual. Certainly, $(\mathbf{p}_{i,k}^{L}, \mathbf{p}_{(i^{\prime },k-1)}^{L})\in \varpi^{\prime}$ is also a valid association obtained through the two-stage feature matching process.

In section \uppercase\expandafter{\romannumeral2}.B, we introduced the compatibility graph to filter out unreliable associations, mitigating issues that could degrade optimization. Each correspondence is associated with a consistent voting score indicating its level of reliability. Consequently, we understand the potential positive contribution of each association. Leveraging the voting results from the two-stage feature matching process, we assign higher weights to these more reliable associations. The process of designing custom weights can be formulated as follows:
\begin{equation}
	W_{i}=\begin{cases}\alpha \cdot \frac{o_{i}-o_{min}}{o_{max}-o_{min}} ,i\in \left [ 0,\lambda \left | \mathbb{O}^{\prime }  \right |  \right ]
		\\
		1,i\in\left (  \lambda \left | \mathbb{O}^{\prime }  \right |, \left | \mathbb{O}^{\prime }  \right |\right ] 
	\end{cases}
	\label{equation::weight}
\end{equation}

The associations in the top $\lambda\%$ of the ordered set $\mathbb{O}^{\prime }$ receive custom optimization weights. $o_{min}$ and $o_{max}$ represent the minimum and maximum scores in the set $\mathbb{O}^{\prime }$, and $\alpha$ is a scale factor. Additionally, the weights for the remaining associations remain unaltered.

Finally, the pose is estimated by minimizing the total of weighted residual terms:
\begin{equation}
	\underset{T_{k}^{k-1}}{\min } \sum {W_{i}f_{e}^{L}(v_{i})}+\sum {W_{j}f_{s}^{L}(v_{j})}
	\label{equation::cosfunc}
\end{equation}

With the aid of cost function (\ref{equation::cosfunc}), we can determine the pose $T_{k}^{k-1}$ using the Levenberg-Marquardt method\cite{hartley2003multiple} as follows:
\begin{equation}
	T_{k}^{k-1}\gets T_{k}^{k-1}-(J^{T}J+\lambda diag(J^{T}J))^{-1}J^{T}f
	\label{equation::update}
\end{equation}

In the optimization process, where $\lambda$ is the Lagrange multiplier, $J$ is the Jacobian matrix, and $f$ is the residual vector term, equation (\ref{equation::update}) iterates to minimize the cost function (\ref{equation::cosfunc}) and obtain the transformation $T_{k}^{k-1}$.
The residual terms linked to these highly reliable correspondences, with increased weights, play a predominant role during optimization. As a result, the gradient descent direction and parameter updates are predominantly guided by these high-quality associations, leading to a more accurate and closer convergence to the ground truth in pose estimation.

\subsection{Lightweight LiDAR Mapping}
The mapping module, typically the back-end, handles precise global pose estimation but at a lower frequency. However, we now present a streamlined mapping module that balances accuracy and efficiency.

During the two-stage feature matching in the mapping module, a KD-tree identifies the neighborhood set $C$, consisting of the five nearest objects to each feature $\mathbf{p}{i}$ from the mapped cloud. In the second graph-based stage, we assign the centroid $\mathbf{p} {i^{\prime },m}$ of $C$ as the temporary corresponding object of feature $\mathbf{p}{i}$, establishing feature correspondence relationships $(\mathbf{p}{i}, \mathbf{\bar{p} } _{i^{\prime },m})$. These relationships are utilized to construct the compatibility graph and perform voting.
While the compatibility graph is used to remove unreliable correspondences during mapping, the voting results are not utilized to weigh the importance of each association.


In the optimization process of the mapping module, two types of residual terms are constructed: a point-to-line residual term and a point-to-plane residual term. The residual terms $f_{e}^{M}(v_{i})$ and $f_{s}^{M}(v_{j})$ are formulated using the same equations as (\ref{equation::residual_edge}) and (\ref{equation::residual_planar}), respectively, which aligns with the LOAM-based solution\cite{zhang2014loam}. The cost function is defined as:
\begin{equation}
	\underset{T_{k}^{M}}{\min } \sum {f_{e}^{M}(v_{i})}+\sum {f_{s}^{M}(v_{j})}
	\label{equation::cosfunc_map}
\end{equation}

In this optimization process, we estimate the global pose $T_{k}^{M}$ using the Levenberg-Marquardt method\cite{hartley2003multiple} without any preferential treatment among the residual terms.

Thanks to the removal of unreliable associations and the guidance from the two-stage feature matching, odometry poses converge quickly. This provides a more accurate initial transformation estimation for the mapping module, resulting in a faster and more precise global pose calculation. The corresponding results are presented in section \uppercase\expandafter{\romannumeral4}. B.
\begin{figure}[h]
	\centering
	\includegraphics[width=0.4\textwidth]{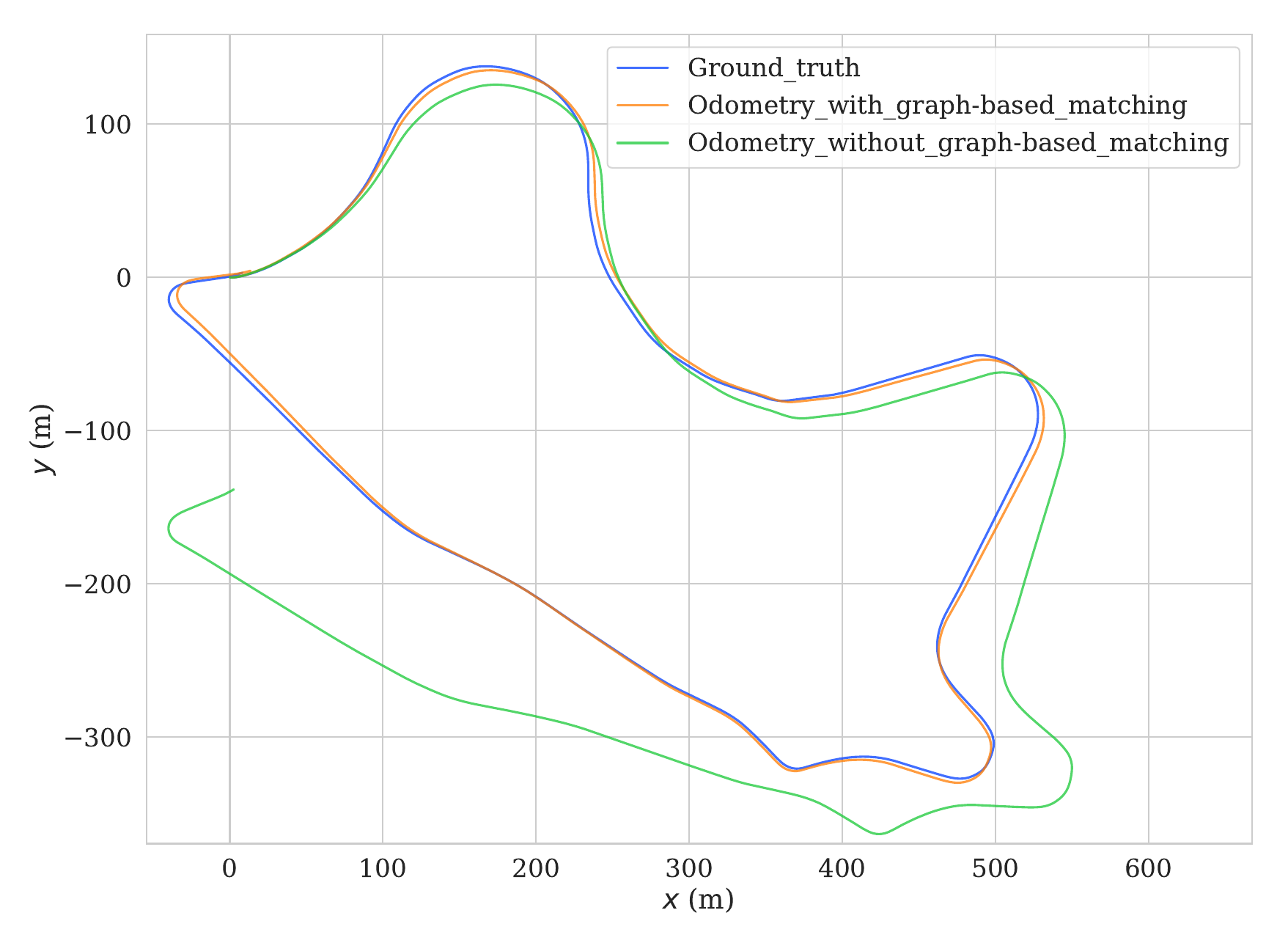}
	\caption{Trajectories estimated by the different front-end odometry modules for the KITTI 09 sequence.}
	\label{fig:odometry_test}
\end{figure}
\section{Experiments}
\subsection{Experiment Setup}
To evaluate the performance of our Light-LOAM SLAM system, we conducted a series of experiments using both the KITTI odometry dataset \cite{kitti} and real-world environments. 
Comprehensive validations, including ablation studies and accuracy assessments, are performed on the KITTI dataset. 
The experiments are conducted on a laptop with an octa-core 3.3GHz processor and 16GB of memory. Real-world testing is carried out on a UAV equipped with an Ouster OS1-32-U LiDAR sensor and a DJI Manifold-2G companion computer featuring an Arm Cortex-A57 CPU and 8GB of memory.

To quantitatively evaluate the accuracy of our SLAM system and facilitate comparison with other approaches, we employed the Absolute Trajectory Error (ATE) metric \cite{ate}. The ATE metric measures the disparity between the estimated poses generated by our system and the ground-truth pose values.

\begin{table*}[]
	\centering
	\caption{RMSE ATE on Public KITTI Odomtery Dataset(meter)}
	\label{tab:outcome_comparison}
	\resizebox{0.98\textwidth}{!}{%
		\begin{tabular}{ccccccccccccc}
			\hline
			Sequence & 00 & 01 & 02 & 03 & 04 & 05 & 06 & 07 & 08 & 09 & 10 & Average \\ \hline
			A-LOAM & 2.624 & 19.795 & 117.323 & 0.851 & \textbf{0.406} & 2.561 & 0.762 & 0.590 & 3.569 & \textbf{1.451} & 1.301 & 13.748 \\
			LeGO-LOAM & 6.339 & 199.682 & 58.276 & 0.991 & 0.549 & 2.586 & 0.960 & 1.110 & 3.993 & 2.183 & 2.397 & 25.370 \\
			HDL-Graph-SLAM & 6.213 & 419.680 & 19.872 & 1.098 & 27.442 & 3.052 & 1.321 & 0.662 & 4.304 & 3.008 & 2.937 & 44.508 \\
			Ours & \textbf{2.315} & \textbf{18.854} & \textbf{10.528} & \textbf{0.849} & 0.411 & \textbf{2.190} & \textbf{0.659} & \textbf{0.587} & \textbf{3.473} & 1.482 & \textbf{1.104} & \textbf{3.859} \\ \hline
		\end{tabular}%
	}

\end{table*}
\subsection{Experiment on KITTI}
\begin{table*}[]
	\centering
	\caption{The Comparison of RMSE ATE Estimated by Front-end Odometry Module(meter)}
	\label{tab:odometry_test}
	\resizebox{0.98\textwidth}{!}{%
		\begin{threeparttable}
			\begin{tabular}{ccccccccccccc}
				\hline
				Sequence & 00 & 01 & 02 & 03 & 04 & 05 & 06 & 07 & 08 & 09 & 10 & Average \\ \hline
				Odomtery(a) & 41.661 & 59.887 & 97.972 & 3.773 & \textbf{1.263} & 24.339 & \textbf{1.941} & 4.918 & 46.183 & 40.706 & 6.667 & 29.937 \\
				Odomtery(b) & \textbf{11.25} & \textbf{29.381} & \textbf{21.238} & \textbf{1.926} & 1.553 & \textbf{5.056} & 2.635 & \textbf{2.72} & \textbf{9.134} & \textbf{3.596} & \textbf{3.407} & \textbf{8.354} \\ \hline
			\end{tabular}%
		\end{threeparttable}
	}
\end{table*}

In this section, we conducted two types of experiments: ablation studies and validations of our Light-LOAM system. To evaluate its performance, we compared Light-LOAM with state-of-the-art LiDAR-based systems, including LOAM\cite{zhang2014loam}, LeGO-LOAM\cite{shan2018lego}, and HDL-Graph-SLAM\cite{hdl-slam}. Notably, for comparison, we used A-LOAM\footnote{\url{https://github.com/HKUST-Aerial-Robotics/A-LOAM}}, an advanced implementation of the LOAM system, instead of the older LOAM version.
\begin{table}[]
	\centering
	\caption{Ablation Study of Light-LOAM}
	\label{tab::balation}
	\resizebox{0.95\columnwidth}{!}{%
		\begin{threeparttable}
			
			\begin{tabular}{ccccc}
				\hline
				Method            & Ours(a) & Ours(b) & Ours(a+b)     \\ \hline
				Average of ATE(m) & 12.138          & 4.018   & \textbf{3.859} \\ \hline
			\end{tabular}
			\begin{tablenotes}
				\footnotesize
				\item  Note that Our(a) refers to our Light-LOAM version only equipped with the non-conspicuous feature selection strategy. Our(b) refers to the Light-LOAM version only equipped with the two-stage feature matching module. And Our(a+b) refers to the version with all modules.
			\end{tablenotes}
		\end{threeparttable}
	}
\end{table}
\begin{table}[]
	\centering
	\caption{AVERAGE RUNTIME OF EACH MODULE FOR ONE SCAN(ms)}
	\label{tab:consuming_time}
	\resizebox{0.95\columnwidth}{!}{%
		\begin{tabular}{cccc}
			\hline
			Method         & Pre-Pocessing & Odometry      & Mapping       \\ \hline
			Ours           & 28.7          & 60.1          & \textbf{61.5} \\
			A-LOAM         & \textbf{19.2} & 45.0          & 89.9          \\
			LeGO-LOAM      & 38.0          & \textbf{20.2} & 82.0          \\
			HDL-Graph-SLAM & 26.4          & 21.7          & 239.7         \\ \hline
		\end{tabular}%
	}
\end{table}
\subsubsection{Abalation Study}
To assess the impact of our advanced algorithm on the Light-LOAM SLAM system, we conducted ablation studies on the inconspicuous feature selection algorithm and the graph-based two-stage feature matching method.

When evaluating the non-conspicuous feature selection method, Our(a) can also be viewed as the LOAM system with the inconspicuous feature selection. The result presented in Table \ref{tab::balation} clearly demonstrates that the non-conspicuous feature selection method improves the pose estimation accuracy of the Light-LOAM system.
This suggests that the inconspicuous selection strategy enhances the ability of our Light-LOAM to select more reliable feature samples.

\begin{figure}[h]
	\centering
	\includegraphics[width=0.4\textwidth]{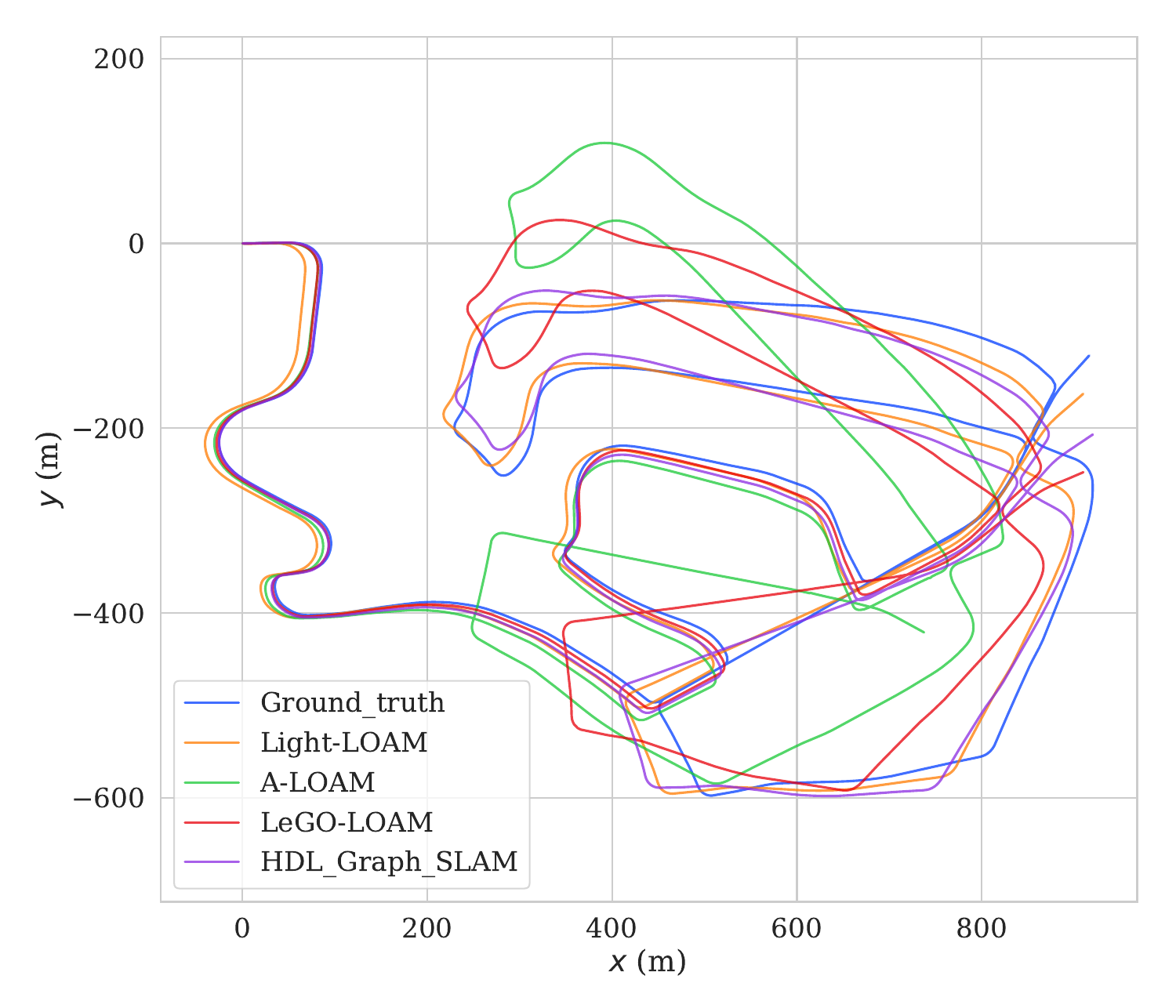}
	\caption{Ground truth and estimated trajectories for the KITTI 02 sequence. }
	\label{fig:kitti02}
\end{figure}
\begin{figure*}[thp]
	\centering
	\includegraphics[width=0.96\textwidth]{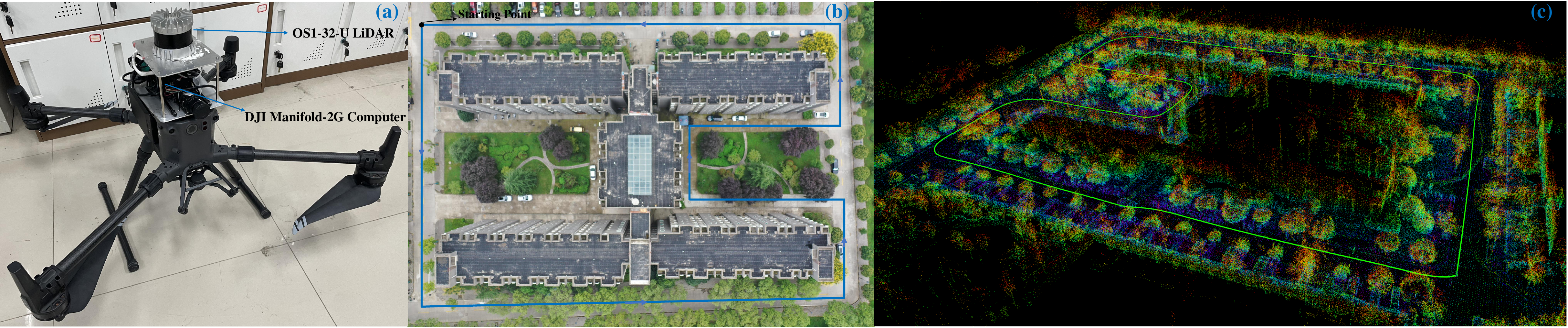}
	\caption{Light-LOAM system in a real-world environment. (a) DJI M300 UAV for experiment. (b) The campus building. (c) Mapping result of Light-LOAM.}
	\label{fig:real_experiment_of_Light-LOAM_system}
\end{figure*}

Our(b) represents the Light-LOAM version with only the two-stage graph-based feature matching module, and its localization result in Table \ref{tab::balation} indicates a substantial improvement in the accuracy of pose estimation. Additionally, we assess the effectiveness of our graph-based matching method by comparing it with the LOAM system in the front-end odometry stage. In this ablation experiment, there are no differences between Light-LOAM and LOAM, except for the inclusion of our graph-based feature matching method.
The results in Table \ref{tab:odometry_test} emphasize the improved precision of odometry's estimated poses when utilizing the graph-based feature matching method. 
Furthermore, the trajectory results for sequence 09, as shown in Fig. \ref{fig:odometry_test}, also demonstrate that the path estimated by the front-end is closer to the ground truth with the assistance of the graph-based feature matching mechanism. Our graph-based feature matching approach excels at filtering out incorrect correspondence relationships and significantly expediting the convergence of pose optimization. In essence, this method shows promise in achieving accurate pose estimation with a limited number of point-pairs, guided by our graph-based feature matching strategy.

\subsubsection{Validation of Light-LOAM}
In this subsection, we evaluate the performance of our Light-LOAM SLAM system in terms of accuracy and efficiency. We start by comparing it with state-of-the-art solutions, as shown in Table \ref{tab:outcome_comparison}. The results clearly indicate that our Light-LOAM system consistently outperforms others in terms of average performance. Even in challenging scenarios, such as sequences 01 and 02 of the KITTI dataset, Light-LOAM maintains its strong performance. For a visual representation of these results, refer to Fig. \ref{fig:kitti02}, where our Light-LOAM exhibits the lowest trajectory errors. These findings underscore the robustness and effectiveness of Light-LOAM, especially in environments where other SLAM algorithms struggle to provide accurate pose estimations.

In addition to assessing accuracy, we have also evaluated the efficiency of our Light-LOAM system. Table \ref{tab:consuming_time} provides an overview of the average runtime results. Notably, our back-end operates with the lowest computational demands, and the mapping module within our SLAM system is lightweight. This improvement in efficiency is attributed to the more precise initial poses provided by consistency-guided odometry. These findings highlight our system's ability to efficiently maintain a global map. Light-LOAM can take a well-balanced compromise between computational cost and localization accuracy.

\subsection{Experiment on Realworld Environment}
\begin{figure}[h]
	\centering
	\includegraphics[width=0.4\textwidth]{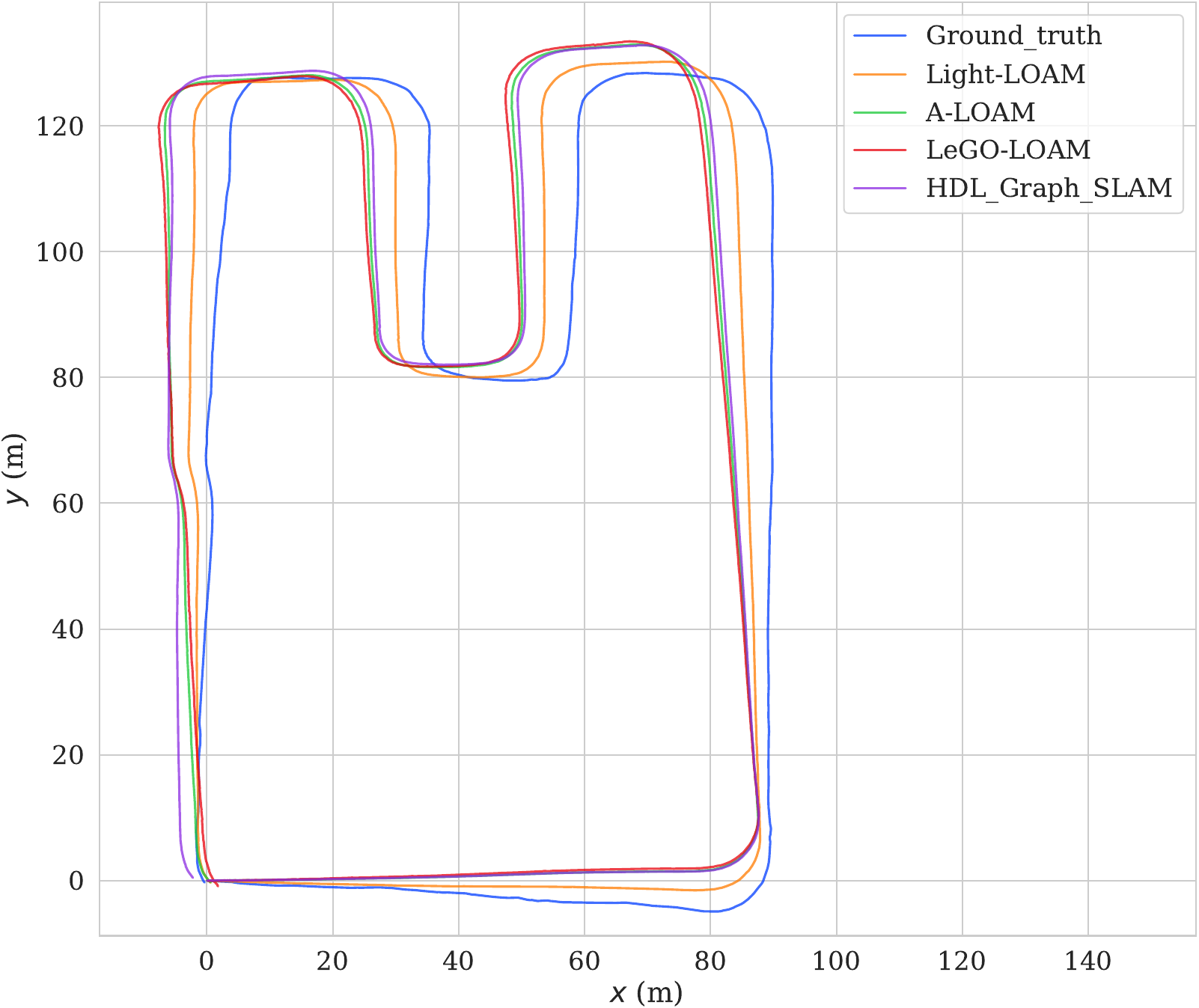}
	\caption{Trajectories estimated by different systems in the real-world environment.}
	\label{fig:realworld_result}
\end{figure}

In this section, we assess the localization and mapping performance of Light-LOAM in a real-world dormitory building using a DJI M300 UAV, depicted in Fig. \ref{fig:real_experiment_of_Light-LOAM_system}(a).
The UAV circled the building at 1.2 m/s and returned to the starting point to check for closed-loop trajectory formation.
For comparison purposes, we also conducted the same localization tasks using state-of-the-art solutions on the same device. We utilized the results from the RTK-GPS(Real-Time Kinematic Global Positioning System) module on the DJI M300 as our ground truth for comparison. The estimated trajectories are visualized in Fig. \ref{fig:realworld_result}, and the results are summarized in Table \ref{tab:realworld_ate}. Notably, the Absolute Trajectory Error (ATE) of Light-LOAM is 0.991, which is lower than its competitors. The outcome of the localization and mapping performance is presented in Fig. \ref{fig:real_experiment_of_Light-LOAM_system}, demonstrating that our mapping results exhibit high accuracy and are capable of completing loop closures without the need for additional loop closure techniques in real-world environments.
These results underscore the effectiveness and precision of our Light-LOAM system in real-world scenarios.
\begin{table}[]
	\centering
	\caption{RMSE ATE of Realworld Scenario}
	\label{tab:realworld_ate}
	\resizebox{0.95\columnwidth}{!}{%
		\begin{tabular}{ccccc}
			\hline
			Method & A-LOAM & LeGO-LOAM & HDL-Graph-SLAM & Ours \\ \hline
			ATE(m) & 1.246 & 1.323 & 1.2589 & \textbf{0.991} \\ \hline
		\end{tabular}%
	}
\end{table}

\section{Conclusion and Future Work}
In this letter, we introduce a lightweight LiDAR SLAM system, Light-LOAM, that employs graph-based matching techniques for efficient and accurate pose estimation. Departing from traditional LOAM-based approaches, we propose a non-conspicuous feature extraction strategy for obtaining stable features. Our graph-based two-stage feature matching method assesses the consistency of associations, filtering unreliable correspondences. The consistency-guided odometry module provides reliable initial pose estimations, and the lightweight mapping module completes localization and mapping tasks.
Experiments on the KITTI odometry dataset and in real-world scenarios show Light-LOAM outperforming the state-of-the-art solutions in accuracy and efficiency. The consistency graph effectively filters out outlier associations, enabling high-quality pose optimization with a limited number of feature samples. Future work involves integrating IMU data to mitigate point cloud distortion and designing a more robust graph-based feature matching method for enhanced reliability in data associations, leading to more accurate pose estimations.
\bibliographystyle{unsrt}
\bibliography{bibtex/bib/cit}
\end{document}